\documentclass[11pt]{article}

\usepackage[final]{acl}

\usepackage{times}
\usepackage{latexsym}
\usepackage{url}
\usepackage{capt-of}
\usepackage[utf8]{inputenc} 
\usepackage[T1]{fontenc}    
\usepackage{url}            
\usepackage{booktabs}       
\usepackage[table]{xcolor} 
\usepackage{amsfonts}       
\usepackage{nicefrac}       
\usepackage{microtype}      
\usepackage{amsmath}
\usepackage{multirow}
\usepackage{subcaption}
\usepackage{pifont}
\usepackage{algorithm}
\usepackage{bbm}

\usepackage{textcomp}
\usepackage{stfloats}
\usepackage{mathabx}
\usepackage{amsthm}
\usepackage{enumitem}
\setlist{leftmargin=5mm}
\usepackage{wrapfig}
\usepackage{makecell}
\usepackage{wrapfig}
\usepackage{lipsum}
\usepackage{amsmath, amssymb, graphicx, xcolor, xspace}
\usepackage{booktabs}
\usepackage{algpseudocode}
\usepackage{bbm}
\usepackage{tcolorbox}  
\usepackage{float}

\definecolor{mygray}{gray}{.9}
\newtcolorbox{mybox}[1][]{
    title=#1,
    fonttitle=\small,
    fontupper=\small,
    left=2mm,
    right=2mm,
    top=1mm,
    bottom=0mm,
}
\usepackage{tikz}
\makeatletter
\newcommand{\DrawLine}{%
  \begin{tikzpicture}
  \path[use as bounding box] (0,0) -- (\linewidth,0);
  \draw[color=black,dashed,dash phase=2pt]
        (0-\kvtcb@leftlower-\kvtcb@boxsep,0)--
        (\linewidth+\kvtcb@rightlower+\kvtcb@boxsep,0);
  \end{tikzpicture}%
  }
\definecolor{violet}{RGB}{138, 43, 226}

\makeatother

\definecolor{citecol}{HTML}{2DDC0E}
\definecolor{tableofcontent}{HTML}{E63E15}
\definecolor{urlcol}{HTML}{2470D8}
\definecolor{myorange}{RGB}{2, 142, 2}

\NewDocumentCommand{\shibo}{ mO{} }
{\textcolor{pink}{\textsuperscript{\textit{Shibo}}\textsf{\textbf{\small[#1]}}}}
\newcommand\ours{\textsc{ArrowGEV}\xspace}
\def\huggingface{\raisebox{-1.5pt}{\includegraphics[height=1.05em]{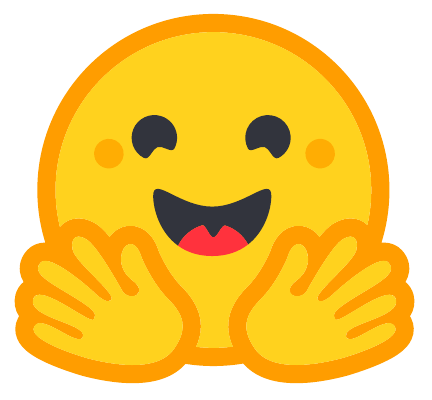}}}
\def\github{\raisebox{-1.5pt}{\includegraphics[height=1.0em]{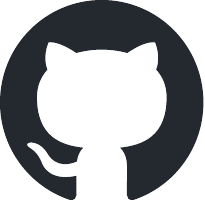}}}
\usepackage[T1]{fontenc}

\usepackage[utf8]{inputenc}

\usepackage{microtype}

\usepackage{inconsolata}

\usepackage{graphicx}

\title{ArrowGEV: Grounding Events in Video via \\ Learning the Arrow of Time}

\author{
\github \enspace \href{https://github.com/Yu-Fangxu/ArrowGEV}{Code} \quad \huggingface \enspace \href{https://huggingface.co/ParadiseYu/ArrowGEV-7B}{Model} \quad \huggingface \enspace \href{https://huggingface.co/datasets/ParadiseYu/ArrowGEV-Data}{Data} \\ 
    \textbf{Fangxu Yu\textsuperscript{1}\thanks{\ This work was done during the internship at Pattern Recognition Center, WeChat AI, Tencent Inc, China.}}, \textbf{Ziyao Lu\textsuperscript{2}\thanks{ \ Corresponding author.}}, \textbf{Liqiang Niu\textsuperscript{2}}, \textbf{Fandong Meng\textsuperscript{2}}, 
    \textbf{Jie Zhou\textsuperscript{2}} \\
    \textsuperscript{1}School of Artificial Intelligence, Nanjing University, China \\
    \textsuperscript{2}Pattern Recognition Center, WeChat AI, Tencent Inc, China \\
    {\tt yufx@smail.nju.edu.cn \quad ziyaolu@tencent.com} \\
}

\begin{document}
\maketitle
\begin{abstract}
Grounding events in videos serves as a fundamental capability in video analysis. While Vision Language Models (VLMs) are increasingly employed for this task, existing approaches predominantly train models to associate events with timestamps in the forward video only. This paradigm hinders VLMs from capturing the inherent temporal structure and directionality of events, thereby limiting robustness and generalization. 
To address this limitation, inspired by the \textit{arrow of time} in physics, which characterizes the intrinsic directionality of temporal processes, we propose \ours, a reinforcement learning framework that explicitly models temporal directionality in events to improve both event grounding and temporal directionality understanding in VLMs. Specifically, we categorize events into time-sensitive (e.g., putting down a bag) and time-insensitive (e.g., holding a towel in the left hand). The former denote events whose reversal substantially alters their meaning, while the latter remain semantically unchanged under reversal. For time-sensitive events, \ours introduces a reward that encourages VLMs to discriminate between forward and backward videos, whereas for time-insensitive events, it enforces consistent grounding across both directions. Extensive experiments demonstrate that \ours not only improves grounding precision and temporal directionality recognition, but also enhances general video understanding and reasoning ability.
\end{abstract}

\begin{figure*}[t]   
	
\centering

    \includegraphics[width=1\textwidth]{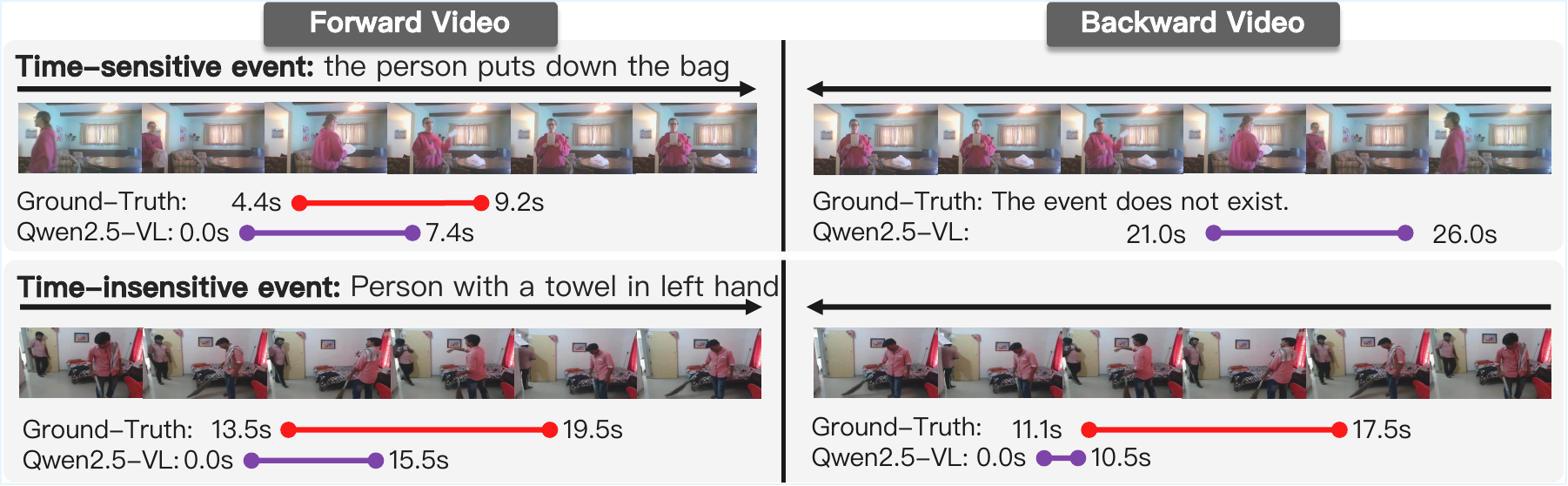}
    \captionof{figure}{Two examples of Qwen2.5-VL-7B predicting event timestamps in forward and backward videos. In the top row, reversing the video changes the event semantics, while in the bottom row, the event remains invariant. The model partially localizes events in forward videos but fails to recognize event absence in the first reversed case and cannot robustly localize the event in the second backward video.}
    \label{fig: pilot_case}
\vspace{-10pt}
\end{figure*}

\section{Introduction}

\label{sec:intro}

Grounding events in videos (GEV) is the task of localizing a specific timestamp in untrimmed videos described by natural language events~\citep{anne2017didemo, lin2023univtg}. As a fundamental ability in video analysis, GEV is crucial for fine-grained video analysis~\citep{luo2025videoautoarena, yang2025streamagent}, video content retrieval~\citep{aslandogan2002techniques, gabeur2020multi}, dense video caption~\citep{iashin2020multi, seo2022end}, and video generation~\citep{tan2024video, yang2025scalingnoise, menapace2024snap}. 

To tackle this challenge, early approaches relied on handcrafted architectures and video-query feature-matching strategies~\citep{hou2022cone, pan2023scanning}, but they suffer from constraints of predefined video snippets, suboptimal video-text features, and poor cross-task generalization. Recent work has shifted to end-to-end Vision Language Models (VLMs)~\citep{wang2025internvl3, team2025kwai, coreteam2025mimovltechnicalreport}, which directly process videos and queries while retaining generalizability either via large-scale timestamp annotation training~\citep{huang2024vtimellm, li2024groundinggpt, zeng2025timesuite}, integration of textual timestamp tokens/embeddings~\citep{guo2025vtg}, or adaptation to event grounding via video segmentation~\citep{guo2024trace, wang2024hawkeye}. Despite this progress, existing methods only align events with forward videos, failing to capture the intrinsic temporal structure and directionality of events.

To investigate this limitation, we present a pilot study (Section~\ref{sec: pilot_study}) and case analysis (Figure~\ref{fig: pilot_case}). As shown in Figure~\ref{fig: pilot_case}, VLMs often struggle to recognize that reversing a video can fundamentally change the semantics of the event, mistakenly associating reversed events with their forward counterparts. In contrast, for events unaffected by temporal reversal, models struggle to consistently locate timestamps in both directions.

To address this, we turn to the \textit{Arrow of Time}~\citep{eddington2019nature, layzer1975arrow}, a foundational concept in physics characterizing the intrinsic directionality of temporal processes. This perspective highlights that the semantics of real-world events are inherently tied to their temporal progression: reversing time may either yield a semantically distinct event or preserve the core meaning, depending on the nature of the process. The impact of this temporal directionality on event understanding manifests itself in two distinct event types: time-sensitive events, where reversal fundamentally alters meaning (e.g., "a man picks up a glass" becomes "a man puts down a glass"), and time-insensitive events, whose meaning remains invariant under reversal (e.g., "a ball is on the table" retains the same semantics when time is reversed).

Inspired by \textit{the Arrow of Time}, we propose \ours, a reinforcement learning (RL) framework, aiming to improve event grounding and the understanding of temporal directionality via learning the arrow of time. Unlike specialized architectures~\citep{chen2021end, woo2024let}, our approach leverages RL to optimize temporal precision through a tailored reward signal directly, mitigating the tendency of VLMs to overfit textual timestamps rather than video semantics. At its core, \ours enables the VLM to learn temporal structures by discriminating between time-sensitive and time-insensitive events. We introduce a reward function that encourages distinct localizations for time-sensitive events and their reversed counterparts while enforcing consistent grounding for time-insensitive ones. To further enhance training efficiency, we propose a difficulty-aware strategy that dynamically emphasizes challenging samples through weighted adjustments and a curriculum-based filtering of well-solved examples.

We conduct extensive experiments on three GEV benchmark datasets, and results indicate that VLMs trained with \ours significantly improve the event grounding performance. In addition, \ours substantially improved the VLM's ability to understand temporal structures in event grounding. Finally, \ours also improves the out-of-distribution (OOD) performance on general video understanding and reasoning tasks.

\section{Related Work}
\noindent\textbf{Grounding Events in Videos with VLMs.}
This task localizes specific events within untrimmed videos~\citep{nan2021interventional,wang2018temporal,li2020tea, zhao2017temporal, kulkarni2025avatar, chen2025scaling, yang2025streamagent, tian2025ego, hannan2024rgnet, mu2024snag, kim2024you, oncescu2021queryd}. Although traditional methods rely on task-specific heads, recent Vision Language Models (VLMs) leverage the reasoning capabilities of LLMs for unified temporal understanding \citep{huang2024vtimellm, li2024groundinggpt}. To bridge the modality gap, existing research typically focuses on large-scale supervised fine-tuning with timestamp-based data \citep{zeng2025timesuite}, introducing specialized temporal tokens \citep{hong2024cogvlm2}, or utilizing video segmentation to align structural granularity \citep{huang2024lita, wang2024hawkeye}. Despite these advances, current VLM-based approaches largely overlook the intrinsic temporal directionality of events. In contrast, \ours introduces a principled framework that explicitly models temporal direction, facilitating more robust and physically consistent grounding.

\noindent\textbf{Post-Training for VLMs.} 
Post-training techniques are essential for adapting pre-trained VLMs to complex downstream tasks. Although large-scale instruction tuning has significantly improved performance in models such as LLaVA-OV~\citep{li2024llavaov} and MAmmoTH-VL~\citep{guo2024mammoth}, recent research has been geared toward RL to refine multimodal reasoning \citep{yu2024flow, su2025pixel, meng2025mm, chen2025mvi}. This shift is particularly evident in video understanding, where RL is increasingly employed to enhance event grounding and long-form reasoning \citep{feng2025videor1, wang2025time}. However, existing RL approaches for videos often overlook the inherent temporal structure of events. Our work addresses this gap by explicitly instilling temporal directionality, guiding VLMs toward a more robust and physically grounded understanding of videos.

\noindent\textbf{Time in Video.}
Temporal directionality is a foundational self-supervised signal for video representation learning, typically utilized through shuffle-and-learn or order-prediction tasks \citep{misra2016shuffle, wei2018learningaot, dwibedi2019temporal, yu2025ts}. VLM frameworks often treat backward videos only as negative samples for contrastive alignment or as binary classification targets \citep{xu2021videoclip, price2019retro}. However, recent evidence suggests that VLMs remain insensitive to temporal directionality in complex reasoning tasks \citep{Xue2025SeeingTA, du2024reversed}. Unlike prior work that focuses on events whose semantics change under reversal, we study both time-sensitive and time-invariant events and move beyond coarse classification. We examine temporal directionality under the high-precision demands of event grounding, requiring models not only to detect reversal but also to localize events along the temporal axis.
\begin{figure*}[t]   
	
\centering

\includegraphics[width=2\columnwidth]{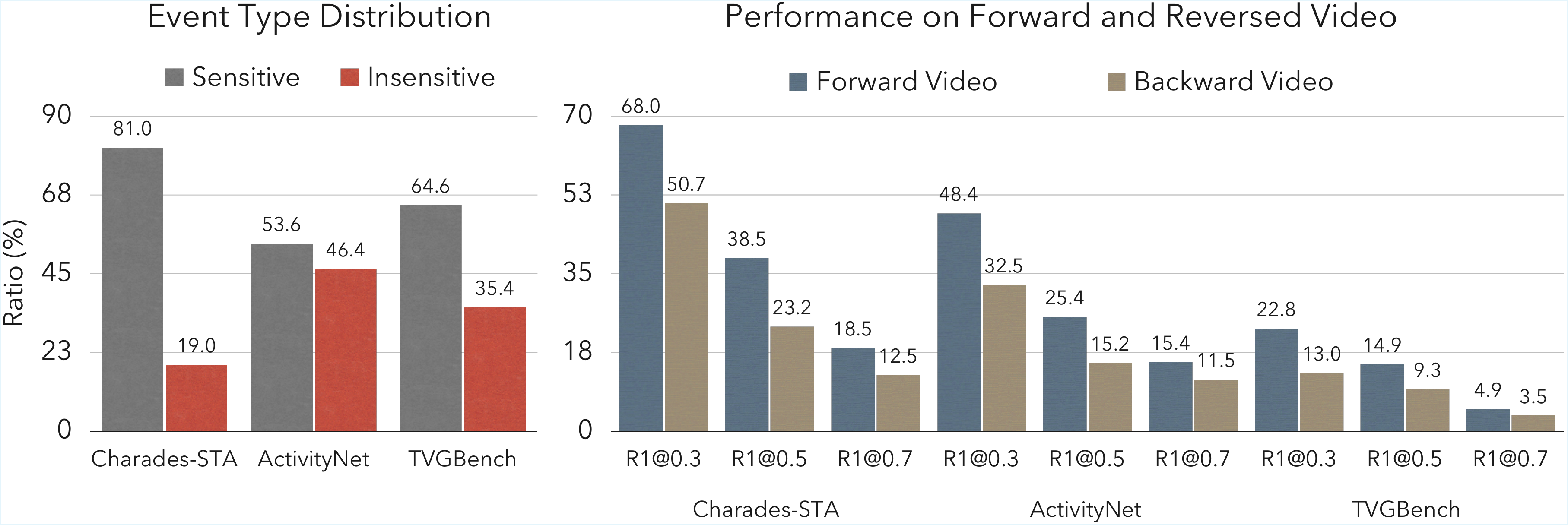}

\caption{Quantitative Analysis of Qwen2.5-VL-7B on GEV Benchmarks. \textbf{Left:} statistics of the portion of time-sensitive and time-insensitive events across three benchmarks. \textbf{Right:} R1@m metrics on the time-sensitive subsets of three benchmarks.}

\label{fig: pilot_study}
\vspace{-10pt}
\end{figure*}

\section{Method}

\subsection{Pilot Study}
\label{sec: pilot_study}
To assess the ability of current Vision Language Models (VLMs) to perceive temporal structure, we conduct a pilot study using the Qwen2.5-7B-VL-Instruct model. Our investigation focuses on time-sensitive events whose semantic meaning is fundamentally altered upon time reversal (e.g., "opening a door" becomes "closing a door"). As illustrated in Figure~\ref{fig: pilot_study}, such events constitute a significant portion of common benchmarks, particularly Charades-STA~\citep{sigurdsson2016Charades}.

We evaluate the VLM by prompting it to localize time-sensitive events in both forward and backward videos. We then measure localization accuracy using Intersection over Union (IoU). For the forward video, we use the original ground truth, while for the backward video, we use a corresponding pseudo-ground truth created by reversing the original event's timestamps.
Ideally, a model with a robust grasp of temporal structure and directionality should recognize that the described event no longer exists in the reversed sequence, resulting in an IoU score close to zero. However, as shown in the right panel of Figure~\ref{fig: pilot_study}, the VLM localizes these time-sensitive events even in backward videos, producing a non-trivial IoU. This reveals a critical failure: the model fails to capture temporal structure, struggling to distinguish the semantic change of events in forward and backward video.
\subsection{Background of GRPO: RL for LLM}
\label{subsec: background}

As a pioneer among open-source R1-style LLMs, DeepSeek-R1~\citep{guo2025deepseek} leverages Group Relative Policy Optimization (GRPO) to train the policy model $\pi_\theta$ (i.e., the LLM) to think before answering, making it particularly suitable for tasks with well-defined answers, such as mathematical reasoning.
In the GRPO framework, given an input question $p$, the LLM samples $G$ candidate responses $o=\{o_1, \dots,o_G\}$, and a reward function $r(\cdot)$ assigns a reward score to each response, yielding $\{r(o_1), \dots, r(o_G)\}$.
GRPO encourages the LLM to generate responses that maximize a weighted sum reward $R(o)$, defined by:
\begin{equation}
\label{eq:ro}
R(o) = \sum_{i=1}^G \frac{\pi_\theta(o_i)}{\pi_{\theta_{\text{old}}}(o_i)} \cdot \underbrace{\frac{r(o_i) - \text{mean}(\{r(o_j)\}_{j=1}^G)}{\text{std}(\{r(o_j)\}_{j=1}^G)}}_{\text{Advantage } A_i}
\end{equation}

where $\pi_\theta(o)$ denotes the probability of LLM generating the response $o$, and $\pi_{\theta_{\mathrm{old}}}$ represents the LLM parameters from a recently optimized state. And the latter term is the Advantage $A_i$ of $i$-th candidate.
To ensure training stability and avoid large deviations from the original language model behavior, the final training objective incorporates a KL-divergence regularization term~\citep{guo2025deepseek}, penalizing divergence between $\pi_\theta$ and $\pi_\mathrm{old}$:
\begin{equation}
\label{eq:grpo}
    \max_{\pi_\theta} \mathbb{E}_{o\sim \pi_{\theta_{\mathrm{old}}}(p)} [
        R(o) - 
        \beta \mathrm{D}_\mathrm{KL}(\pi_\theta \| \pi_\mathrm{ref})
    ] \\
\end{equation}

where $\beta$ is a scaling coefficient.
We omit the clipping operation for simplicity.

\subsection{ArrowGEV}
\noindent\textbf{Formulation.} 
The task of grounding events in videos requires a model $\mathcal{M}$ to map an untrimmed video $\mathcal{V}$ of duration $d$ and a natural language event $q$ to a specific temporal timestamp $\mathcal{T}^{\text{fwd}} = \mathcal{M}(\mathcal{V}, q)$ that accurately aligns with a ground-truth annotation $\mathcal{T}^{\text{gt}}$. As discussed in Section~\ref{sec: pilot_study}, the VLM struggles to capture the temporal structure of events in the video. To overcome this limitation, we leverage the backward video, denoted as $\mathcal{V}'$, to enhance the understanding of temporal structure. Our key insight is that the effect of temporal reversal is not uniform across all events; it depends on the event's intrinsic properties. Therefore, to construct a meaningful learning signal from the backward video $\mathcal{V}'$, we first categorize events according to their temporal nature.

\noindent\textbf{Event Categorization.}
We categorize events into two types:

1) \textit{Time-sensitive events}: Events whose semantics are transformed into a distinct, often opposite, action when time is reversed (e.g., ``opening a door'' becomes ``closing a door''). We expect a model to recognize the changed semantics and reduce the prediction on the same video segment.

2) \textit{Time-insensitive events}: Events whose semantics are preserved under temporal reversal (e.g. ``a car is parked''). We expect a model's prediction to be consistent with respect to time reversal.

Then, we design an RL framework to encourage the VLM to not only accurately localize the timestamps of the events in the forward video, but also to distinguish the existence of time-sensitive events and recognize time-insensitive events in the backward video. After training, \ours enables the VLM to achieve above and learn the temporal structure more robustly to enhance the localization. 

\begin{figure*}[t]   
	
\centering

\includegraphics[width=1.0\textwidth]{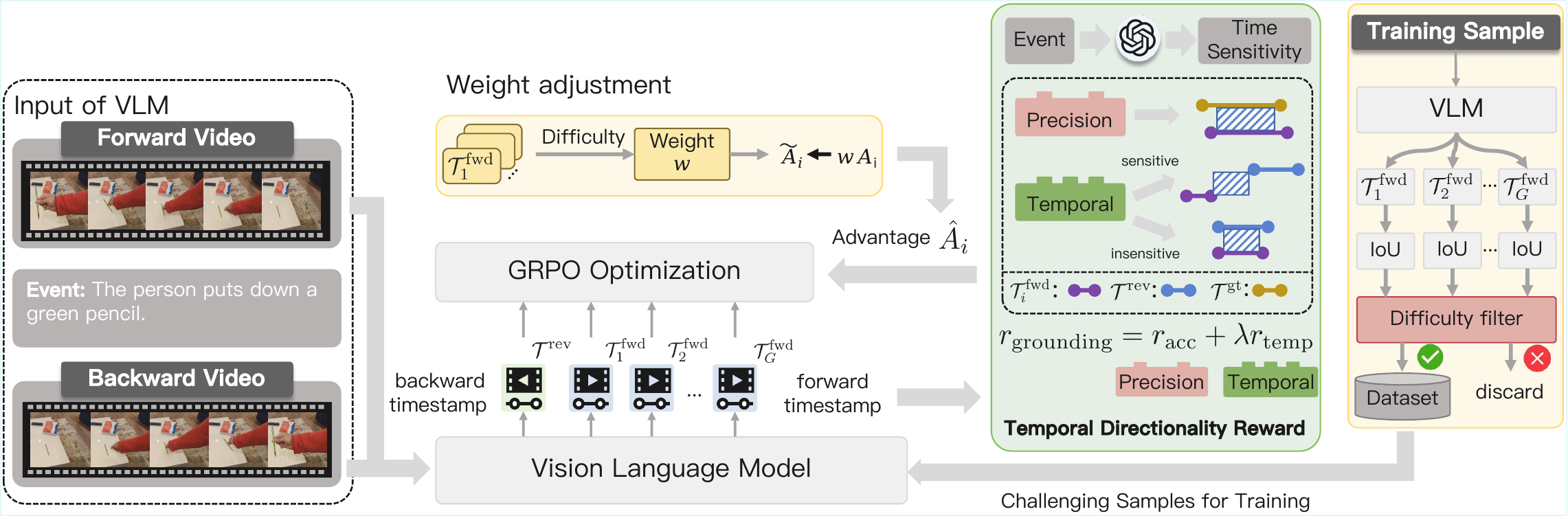}

\caption{Overview of \ours. First, we input the event and both forward and backward videos into the VLM to obtain the predicted timestamps in both directions. Then we calculate the reward based on the category of the event. Having the reward for samples, we use GRPO with difficulty-aware training strategies to optimize the VLM for improved localization accuracy and directionality understanding. }

\label{fig: main_arch}
\end{figure*}
\subsection{Temporal Directionality Reward Modeling}
A reward function for GEV should ideally incentivize a model to localize events while understanding an event's temporal structure. 

To determine the type of each event, we use an LLM to perform reasoning to classify each event $q$ as time-sensitive or time-insensitive with prompt $p$. The model first generates the reasoning process $r \sim P_{\text{LLM}}(\cdot | p, q)$, and then determines the type of event $c(q) \sim P_{\text{LLM}}(\cdot|r, p, q)$. The analysis in Appendix~\ref{sec: classification} verifies the reliability of event categorization. See the prompt in Appendix~\ref{sec: prompt}.

Let $c(q) \in \{ \text{insensitive}, \text{sensitive} \}$ denote the event's category. We formulate the grounding reward $r$ as a linear combination of two components: a localization accuracy reward $r_{\text{acc}}$, which promotes precise event localization in the forward video, and a temporal directionality reward $r_{\text{temp}}$, which enforces temporal understanding. These are balanced by a weighting factor $\lambda$:
\begin{equation}
    r_{\text{grounding}} = r_{\text{acc}} + \lambda r_{\text{temp}}
\end{equation}
The accuracy reward $r_{\text{acc}}$ is defined as timestamp-aware IoU~\citep{wang2025time} with the ground-truth $\mathcal{T}^{\text{gt}}$, which encourages the alignment between start and end time with the ground-truth based on the standard IoU:
\begin{equation}
    \mathrm{tIoU}(\mathcal{T}^{\text{fwd}}, \mathcal{T}^{\text{gt}}) = \text{IoU} \cdot (1 - \frac{|t_s - t'_s|}{d}) \cdot (1 - \frac{|t_e - t'_e|}{d}) 
\end{equation}
\begin{equation}
    r_{\text{acc}} \triangleq \mathrm{tIoU}(\mathcal{T}^{\text{fwd}}, \mathcal{T}^{\text{gt}})
\end{equation}
    
To define the temporal directionality reward, let $\mathcal{T}^{\text{rev}} = \mathcal{M}(\mathcal{V}', q)$ be the VLM's predicted timestamp on the backward video. We define a temporal reversal operator $\mathcal{R}(\mathcal{T}) \triangleq [d - t_e, d - t_s]$ for a timestamp $\mathcal{T}=[t_s, t_e]$ in a video. This allows us to compute a \textbf{directionality score}, $S_c$, measuring the alignment between the reversed prediction and the mirrored forward prediction:
\begin{equation}
    S_c \triangleq \mathrm{tIoU}(\mathcal{T}^{\text{rev}}, \mathcal{R}(\mathcal{T}^{\text{fwd}}))
\end{equation}
The temporal directionality reward $r_{\text{temp}}$ leverages this score to enforce the expected temporal properties of events. For time-insensitive events, a high reward ($r_{\text{temp}} = S_c$) reflects strong overlap between the predicted timestamps in the forward and backward videos, demonstrating consistency in the model’s localization across both directions. In contrast, for time-sensitive events, a low reward is desirable, as it indicates that the model successfully distinguishes the semantic shifts introduced by temporal reversal. Specifically, the reward function ($r_{\text{temp}} = 1-S_c$) penalizes the incorrect localization of time-sensitive events in backward videos, encouraging the generation to indicate the event's nonexistence, rather than an incorrect timestamp.
Together, these cases motivate our final unified reward function:
\begin{equation}
    r_{\text{grounding}} = r_{\text{acc}} + \lambda \cdot
    \begin{cases}
        S_c & \text{if } c(q) = \text{insensitive} \\
        1 - S_c & \text{if } c(q) = \text{sensitive}
    \end{cases}
    \label{eq:final_reward}
\end{equation}
By optimizing Equation~\ref{eq:final_reward}, our model is explicitly incentivized not only to be accurate but also to develop an internal representation that is consistent with the fundamental temporal properties of events.

\noindent\textbf{Reasoning Template Reward.} To facilitate complex temporal reasoning, we adopt a \textit{think-before-act} paradigm, requiring the model to generate intermediate rationales before predicting timestamps. We enforce this via a binary format reward $r_{\text{form}}(o) \in \{0, 1\}$, which validates if the output $o$ strictly adheres to the template: \texttt{<think> ... </think> <answer> <$t_s$ \text{to} $t_e$> </answer>}. 
This structural component is then integrated into the final objective, which combines our format, accuracy, and temporal directionality rewards:
\begin{equation}
    r_{\text{final}} = r_{\text{grounding}} + r_{\mathrm{form}}(o) 
\end{equation}
\subsection{Training}
To train the model, we propose two strategies to address the difficulty-bias issue and improve training efficiency.

\noindent \textbf{Difficulty-Aware Weight Adjustment.} 
During training, samples become progressively easier for the model. This will lead to a difficulty-bias issue. To alleviate this, we propose a weight adjustment that enables the model to focus more on the hard samples. Specifically, we introduce a difficulty coefficient $w_i \propto -\text{tIoU}(\mathcal{T}^{\text{fwd}}, \mathcal{T}^{\text{gt}})$ for $i$-th sample to quantify the difficulty level. In this case, we sample $G$ predictions by the VLM and calculate the difficulty weight:
\begin{equation}
    w_i = \exp\!\left(\frac{1 - \tfrac{1}{G} \sum_{j=1}^{G} \text{tIoU}\!\bigl(\mathcal{T}^{\text{fwd}}_{j}, \mathcal{T}^{\text{gt}}\bigr)}{\tau}\right),
\end{equation}
This coefficient dynamically adjusts sample weights by computing the average tIoU of different responses for $i$-th sample: $\hat{A_i} = w_i A_i$.

\paragraph{Dynamic Curriculum via Difficulty Filtering.}
The efficacy of the above-detailed policy optimization is highly dependent on the quality and challenge of the training data distribution, $\mathcal{D}$. As the policy VLM $\pi_\theta$ improves, a static dataset can be dominated by samples that no longer provide a sufficient learning signal. To counteract this and maintain a challenging training environment, we implement a dynamic curriculum strategy that adaptively refines the data distribution throughout the training. Specifically, we initialize our training set, $\mathcal{D}_0$, with the dataset from~\citep{wang2025time}, which is pre-filtered to emphasize samples of moderate difficulty. To ensure the model is persistently challenged as it learns, we introduce a dynamic difficulty filter at the conclusion of each training epoch $e$. To construct the dataset for the next epoch, $\mathcal{D}_{e+1}$, we evaluate each sample $(\mathcal{V}, q) \in \mathcal{D}_e$ against the current policy $\pi_{\theta_e}$. A sample is deemed "mastered" and is subsequently removed if the policy consistently solves it with high accuracy. Formally, for each $(\mathcal{V}, q)$, we generate a group of $G$ rollout outputs $\{\mathcal{T}_{i}^{fwd}\}_{i=1}^G$. The sample is filtered out if its worst-case performance in the group exceeds a high-performance threshold $\eta$:
\begin{equation}
\small
    \mathcal{D}_{e+1} = \mathcal{D}_e \setminus \left\{ (\mathcal{V},q) \in \mathcal{D}_e \mid \min_{i=1 \dots G} \mathrm{IoU}(\mathcal{T}_{i}^{fwd}, \mathcal{T}^{\text{gt}}) > \eta \right\}
\end{equation}
We set $\eta=0.7$ in our experiments, which is the strictest IoU threshold commonly adopted in prior work~\citep{wang2025time, li2025universal, nguyen2025multi}. This adaptive curriculum ensures that the model continually focuses its capacity on unsolved or challenging problems, thereby maintaining a strong gradient signal and promoting the development of a more robust and generalizable policy.

\section{Experiments}
\subsection{Experimental Settings.}
\textbf{Benchmarks.} We evaluate our model on three GEV benchmarks: Charades-STA~\citep{sigurdsson2016Charades}, ActivityNet~\citep{caba2015activitynet}, and TVGBench~\citep{wang2025time}. To further evaluate the generalization ability, we further compare \ours on the video understanding and reasoning benchmarks, including TempCompass~\citep{liu2024tempcompass}, MVBench~\citep{li2024mvbench}, VSI-Bench~\citep{yang2025thinking}, Video-MMMU~\citep{hu2025video}, MMVU~\citep{zhao2025mmvu}, and VideoMME~\citep{fu2024videomme}.

\noindent\textbf{Implementation Details.} Our methodology is built upon the Qwen2.5-VL-7B-Instruct model~\citep{Qwen2.5-VL}. For computational efficiency, we process videos by sampling frames at 2~FPS. See more training details in Appendix~\ref{sec: training}.

\noindent\textbf{Evaluation Metrics.} Following established protocols~\citep{ren2024timechat, huang2024vtimellm}, we report R1@m at various IoU thresholds. This metric calculates the percentage of test samples where the IoU between the top-ranked predicted temporal segment and the ground truth exceeds a given threshold $m \in \{0.3, 0.5, 0.7\}$. As a complementary metric, we also report the mean IoU (mIoU) averaged across the entire test set. For the video understanding and reasoning tasks, we evaluate performance using standard accuracy. To further quantitatively assess the model's comprehension of temporal directionality, we introduce the \textbf{Temporal Directionality Discrepancy (TDD)} metric. The core idea behind TDD is that a model that truly understands temporal direction should behave differently based on an event's intrinsic time sensitivity. 
It is formally defined as:
\begin{equation}
    \text{TDD(m)} = \frac{\text{R1@m}(\text{fwd})-\text{R1@m}(\text{rev})}{\text{R1@m}(\text{fwd})},
\end{equation}
where $\text{R1@m}(\text{fwd})$ denotes R1@m when predictions align with the ground truth $\mathcal{T}_{gt}$ on the forward video $\mathcal{V}$, and $\text{R1@m}(\text{rev})$ denotes R1@m on the backward video $\mathcal{V}'$ with respect to the mirrored ground truth $\mathcal{R}(\mathcal{T}_{gt})$.
The interpretation of the TDD depends on the event category. For Time-sensitive events, ideal models should accurately localize the forward event ($\text{R1@m}(\text{fwd}) \rightarrow 1$) but recognize its absence upon reversal due to the nonexistence of the event ($\text{R1@m}(\text{rev}) \rightarrow 0$), yielding a TDD approaching 1. For time-insensitive events, models should demonstrate temporal invariance by consistently localizing the event in both directions ($\text{R1@m}(\text{fwd}) \approx \text{R1@m}(\text{rev})$), yielding a TDD approaching 0. This demonstrates that the model correctly recognizes the event's invariance to temporal direction and exhibits strong consistency.

\begin{table*}[t]

\centering
\setlength{\belowcaptionskip}{3pt}%
\resizebox{\textwidth}{!}{
\begin{tabular}{l|ccccc@{}|ccccc@{}|cccc}
\toprule
\multirow{2}{*}{Method} & \multicolumn{4}{c}{Charades-STA} && \multicolumn{4}{c}{ActivityNet} && \multicolumn{4}{c}{TVGBench}\\
& {\fontsize{8.4}{10}\selectfont R1@0.3} & {\fontsize{8.4}{10}\selectfont R1@0.5} & {\fontsize{8.4}{10}\selectfont R1@0.7} & {\fontsize{8.4}{10}\selectfont mIOU} && {\fontsize{8.4}{10}\selectfont R1@0.3} & {\fontsize{8.4}{10}\selectfont R1@0.5} & {\fontsize{8.4}{10}\selectfont R1@0.7} & {\fontsize{8.4}{10}\selectfont mIOU} && {\fontsize{8.4}{10}\selectfont R1@0.3} & {\fontsize{8.4}{10}\selectfont R1@0.5} & {\fontsize{8.4}{10}\selectfont R1@0.7} & {\fontsize{8.4}{10}\selectfont mIOU}\\
\midrule
Gemini-2.5-Flash~\citep{comanici2025gemini} & 47.0 & 21.8 & 7.1 & 31.1 && 33.5 & 20.8 & 14.2 & 25.1 && 31.6 & 21.1 & 5.3 & 21.6 \\
Gemini-2.5-Pro~\citep{comanici2025gemini} & 53.9 & 25.5 & 8.8 & 34.6 && 48.2 & 31.9 & 18.7 & 36.3 && 40.0 & 25.7 & 11.4 & 29.1 \\
GPT-5~\citep{singh2025openai} & 39.7 & 18.3 & 6.2 & 28.4 && 48.2 & 33.0 & 18.8 & 35.9 && 31.2 & 18.8 & 6.3 & 22.3 \\\midrule
\textcolor{gray}{2D-TAN$^*$~\citep{zhang20192DTAN}} &
\textcolor{gray}{57.3} & \textcolor{gray}{45.8} & \textcolor{gray}{27.9} & - && \textcolor{gray}{60.4} & \textcolor{gray}{43.4} & \textcolor{gray}{25.0} & - && - & - & - & - \\
\textcolor{gray}{UniVTG$^*$~\citep{lin2023univtg}} &
\textcolor{gray}{72.6} & \textcolor{gray}{60.2} & \textcolor{gray}{38.6} & - && \textcolor{gray}{56.1} & \textcolor{gray}{43.4} & \textcolor{gray}{24.3} & - && - & - & - & - \\
\textcolor{gray}{SSRN$^*$~\citep{ssrn}} &
- & \textcolor{gray}{65.5} & \textcolor{gray}{42.6} & - && - & \textcolor{gray}{54.5} & \textcolor{gray}{33.2} & - && - & - & - & - \\
\textcolor{gray}{SnAG$^*$~\citep{CVPR2024SnAG}} &
- & \textcolor{gray}{64.6} & \textcolor{gray}{46.2} & - && - & \textcolor{gray}{48.6} & \textcolor{gray}{30.6} & - && - & - & - & - \\
\textcolor{gray}{EaTR$^*$~\citep{jang2023knowing}} &
- & \textcolor{gray}{68.4} & \textcolor{gray}{44.9} & - && - & \textcolor{gray}{58.2} & \textcolor{gray}{37.6} & - && - & - & - & - \\
\textcolor{gray}{HawkEye$^*$~\citep{wang2024hawkeye}} &
\textcolor{gray}{72.5} & \textcolor{gray}{58.3} & \textcolor{gray}{28.8} & - && \textcolor{gray}{55.9} & \textcolor{gray}{34.7} & \textcolor{gray}{17.9} & - && - & - & - & - \\
\textcolor{gray}{TimeSuite$^*$~\citep{zeng2025timesuite}} &
\textcolor{gray}{79.4} & \textcolor{gray}{67.1} & \textcolor{gray}{43.0} & - && - & - & - & - && - & - & - & - \\
\midrule
ChatVTG~\citep{qu2024chatvtg} &
52.7 & 33.0 & 15.9 & - && 40.7 & 22.5 & 9.4 & - && - & - & - & - \\
TimeChat~\citep{ren2024timechat} &
46.7 & 32.2 & 15.7 & 32.2 && 30.2 & 16.9 & 8.2 & 21.8 && 22.4 & 11.9 & 5.3 & - \\
HawkEye~\citep{wang2024hawkeye} &
50.6 & 31.4 & 14.5 & - && 49.1 & 29.3 & 10.7 & - && - & - & - & - \\
VTimeLLM~\citep{huang2024vtimellm} &
51.0 & 27.5 & 11.4 & 31.2 && 44.0 & 27.8 & 14.3 & 30.4 && - & - & - & - \\
TimeSuite~\citep{zeng2025timesuite} &
69.9 & 48.7 & 24.0 & - && - & - & - & - && 31.1 & 18.0 & 8.9 & - \\
Momentor~\citep{qian2024momentor} & 42.9 & 23.0 &  12.4 & 29.3 && 42.6& 26.6 & 11.6 & 28.5 && -& -& -& -\\
VTG-LLM~\citep{guo2025vtg} & 52.0 & 33.8 & 15.7 & - && - & 8.3 & 3.7 & 12.0 && - & - & -& - \\
Time-R1\textsuperscript{\dag}~\citep{wang2025time} & 77.6 & 59.0 & 32.4 & 52.4 && 55.2 & 36.4 & 19.7 & 38.1 && 40.4 & 27.0 & 12.6 & 27.5 \\
TVG-R1\textsuperscript{\dag}~\citep{chen2025datasets} & 60.7 & 36.1 & 13.8 & 39.3 && 53.9 & 33.7 & 17.5 & 37.6 && 32.1 & 18.1 & 9.4 & 23.0 \\
TimeLens-7B~\citep{zhang2025timelens} & 70.7 & 39.8 & 14.5 & 42.3 && 53.5 & 35.2 & 19.7 & 37.7 && 38.3 & 21.7 & 12.2 & 25.7\\
GranAlign~\citep{jeon2026granalign} & 59.1 & 39.6 & 22.7 & 38.0 && 50.3 & 34.0 & 16.5 & 33.1 && - & - & - & -\\
VideoChat-Flash~\citep{li2024videochatflash} &
74.5 & 53.1 & 27.6 & - && - & - & - & - && 32.8 & 19.8 & 10.4 & - \\
TRACE~\citep{guo2024trace} &
- & 40.3 & 19.4 & - && - & \textcolor{gray}{37.7} & \textcolor{gray}{24.0} & - && 37.0 & 25.5 & 14.6 & - \\
\midrule
Qwen-2.5-VL-7B \textsuperscript{\dag} & 59.6 & 38.7 & 16.5 & 38.3 && 33.9 & 21.4 & 13.1 & 25.2 && 25.4 & 16.4 & 8.5 & 17.8\\
\ours-7B & \textbf{78.0} & \textbf{61.6} & \textbf{37.2} & \textbf{54.1} && \textbf{58.5} & \textbf{38.2} & \textbf{20.3} & \textbf{39.9} && \textbf{41.9} & \textbf{29.5} & \textbf{16.0} & \textbf{29.2}\\
\bottomrule
\end{tabular}}
\caption{
Results on GEV benchmarks. The methods marked in \textcolor{gray}{gray$^*$} represent fine-tuning on corresponding benchmarks, while those in black indicate zero-shot settings. \textsuperscript{\dag} denotes that the results are reproduced with the official weights for fair comparison.
}
\label{tab:comp_GEV}
\vspace{-10pt}
\end{table*}

\subsection{Main Results}
Table~\ref{tab:comp_GEV} compares the performance of \ours with state-of-the-art methods. As expected, models trained directly on the target benchmarks generally outperform zero-shot approaches. We further observe that RL-based methods generally achieve competitive or superior performance compared with SFT-based ones, likely because they optimize directly on temporal signals grounded in videos and the strong generalization ability of RL~\citep{chu2025sft, peng2025lmm}. In particular, \ours achieves higher accuracy than the strongest SFT-based baseline on all R1@m metrics. Moreover, compared to other RL-based methods, \ours yields average improvements in R1@m of 2.6\% on Charades-STA, 1.9\% on ActivityNet, and 2.5\% on TVGBench. We attribute these gains to explicit training to understand the temporal structure, which enhances the robustness of the model in localizing events within videos.

\subsection{OOD Generalization}
\begin{figure}[t]   
	
\centering

\includegraphics[width=1.0\columnwidth]{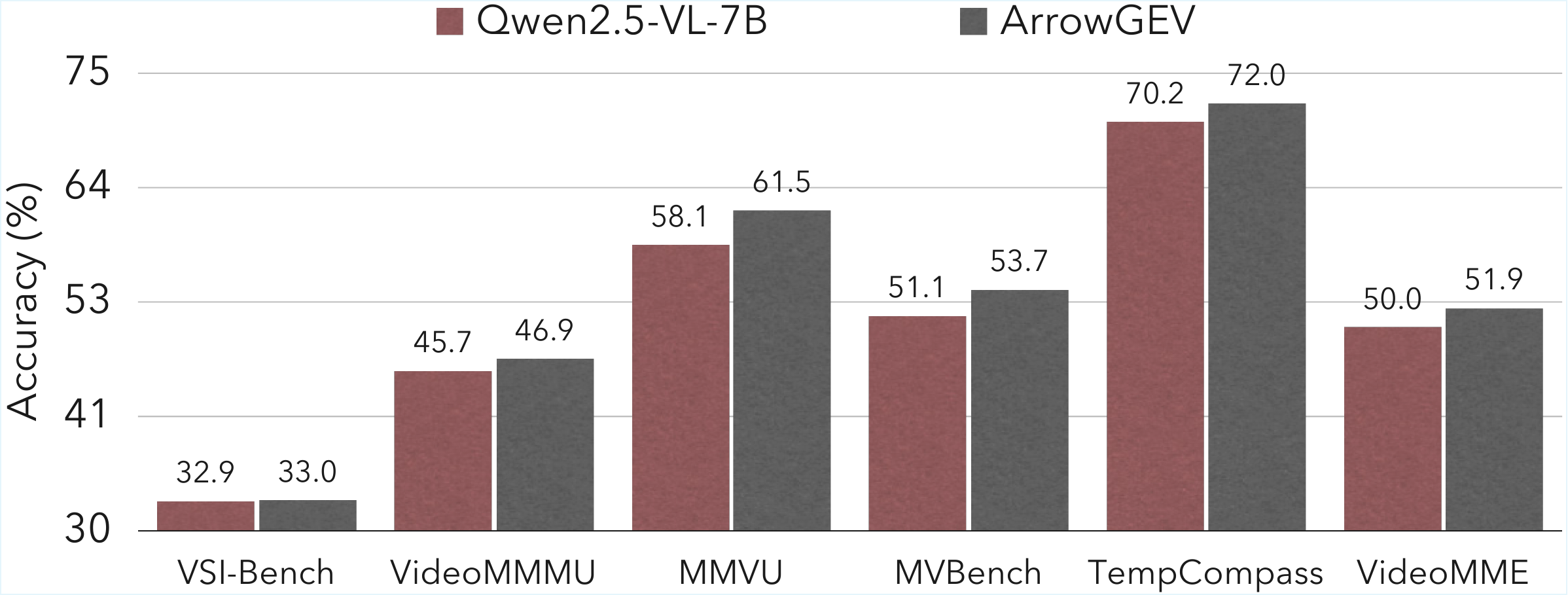}

\caption{OOD results on six general video understanding and reasoning benchmarks.}

\label{fig: performance_OOD}
\vspace{-10pt} 
\end{figure}

Beyond GEV tasks, we further evaluate \ours on general video understanding and reasoning benchmarks. As shown in Figure~\ref{fig: performance_OOD}, \ours yields consistent and significant improvements over the base Qwen2.5-VL-7B model across all benchmarks. These results highlight the versatility of our approach and demonstrate that explicitly modeling the arrow of time enhances OOD generalization across diverse video understanding and reasoning scenarios. We attribute this improvement to \ours's ability to deeply understand temporal structure, which is a core element of general video understanding and reasoning. 

\begin{figure*}[t]   
\centering

\includegraphics[width=2\columnwidth]{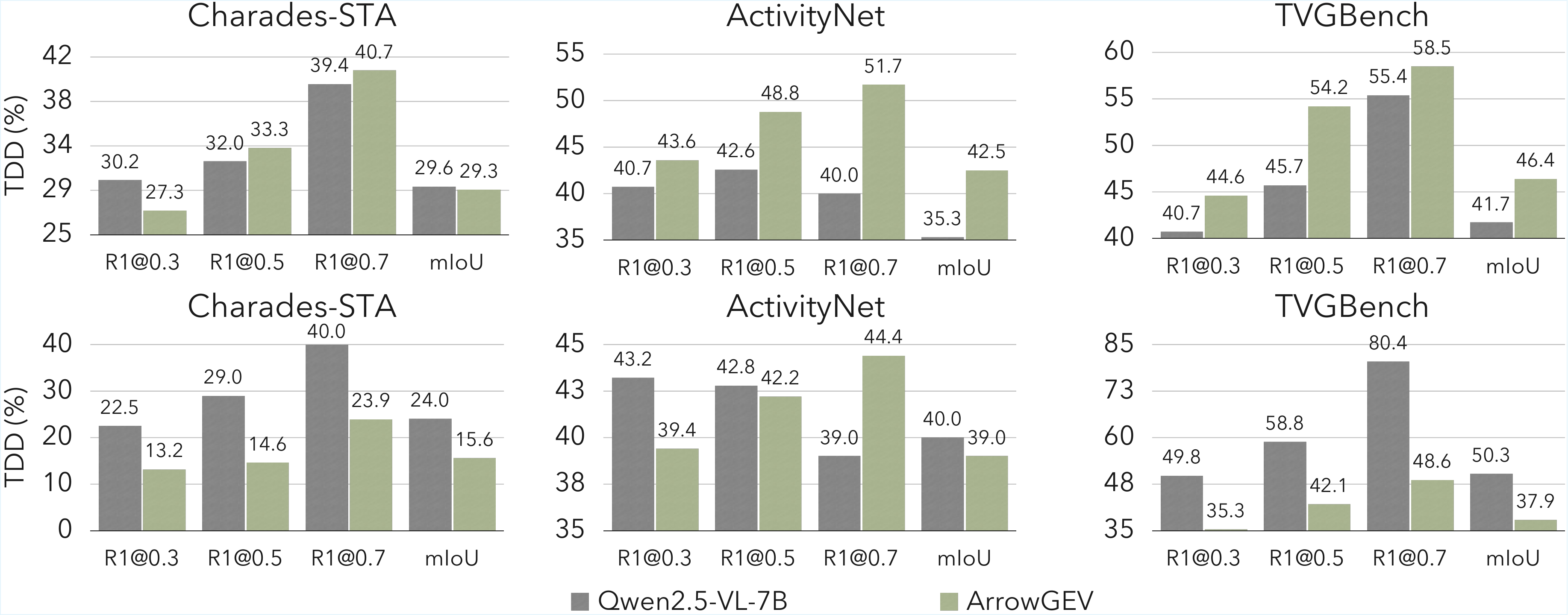}

\caption{Results of the temporal directionality discrepancy (TDD) metric on three benchmarks. The upper row reports results on the time-sensitive subset, where higher values indicate better temporal directionality understanding, while the bottom row shows results on the time-insensitive subset, where lower values are preferable.}

\label{fig: performance_confusion}
\end{figure*}

\subsection{Improvement on Temporal Directionality Understanding}
To quantify temporal directionality, we evaluate \ours on both the time-sensitive subset and the insensitive subset. For the time-sensitive subset, a larger TDD indicates better temporal directionality understanding, as the model is less likely to associate meaning-changing timestamps in the backward video with the event. Conversely, for the insensitive subset, a smaller TDD score reflects better temporal directionality understanding, as it suggests the model consistently recognizes events in both forward and backward videos.

As illustrated in Figure~\ref{fig: performance_confusion}, \ours achieves a substantial improvement over the base model in the time-sensitive and time-insensitive subsets. For time-sensitive subsets (upper row), the TDD score increases markedly, demonstrating that \ours successfully discriminates semantics that diverge under temporal reversal. Conversely, on the time-insensitive subset (bottom row), \ours achieves a significant reduction in TDD. This reduction confirms that the model maintains consistent localization across temporal flips, ensuring robust performance regardless of event directionality. These results indicate that \ours instills a deeper and more robust understanding of temporal directionality in VLMs.
\begin{table*}[t]
\centering
\setlength{\belowcaptionskip}{3pt}%
\resizebox{\textwidth}{!}{
\begin{tabular}{l | cccc | cccc | cccc} 
\toprule
\textbf{Method} & \multicolumn{4}{c|}{\textbf{Charades-STA}} & \multicolumn{4}{c|}{\textbf{ActivityNet}} & \multicolumn{4}{c}{\textbf{TVGBench}} \\
\cmidrule(r){2-5} \cmidrule(lr){6-9} \cmidrule(l){10-13}
 & R1@0.3 & R1@0.5 & R1@0.7 & mIOU & R1@0.3 & R1@0.5 & R1@0.7 & mIOU & R1@0.3 & R1@0.5 & R1@0.7 & mIOU \\
\midrule
Qwen-2.5-VL-7B & 59.6 & 38.7 & 16.5 & 38.3 & 33.9 & 21.4 & 13.1 & 25.2 & 25.4 & 16.4 & 8.5 & 17.8\\
\midrule
\qquad + GRPO & 74.0 & 56.1 & 30.8 & 50.3 & 56.1 & 36.0 & 19.1 & 38.3 & 38.7 & 26.2 & 15.2 & 27.3\\
\qquad + Weight Adj. & 76.6 & 58.9 & 32.5 & 52.1 & 55.4 & 36.2 & 19.6 & 37.9 & 40.3 & 26.7 & 14.7 & 27.6\\
\qquad + Filtering  & 75.2 & 58.6 & 35.1 & 51.9 & 56.1 & 37.1 & 19.7 & 38.7 & 41.0 & 28.5 & 14.1 & 28.0 \\
\qquad + Temporal Reward & \textbf{78.0} & \textbf{61.6} & \textbf{37.2} & \textbf{54.1} & \textbf{58.5} & \textbf{38.2} & \textbf{20.3} & \textbf{39.9} & \textbf{41.9} & \textbf{29.5} & \textbf{16.0} & \textbf{29.2} \\
\bottomrule
\end{tabular}}
\caption{
Ablation results on three GEV benchmarks.
}
\label{tab:comp_GEV_ablation} 
\vspace{-10pt}
\end{table*}
\subsection{Ablation Study}
We conduct ablation studies to analyze the impact of individual components in \ours. Table~\ref{tab:comp_GEV_ablation} summarizes component-wise ablations:
1) Incorporating the temporal reward substantially enhances robustness in event grounding. Adding the temporal reward yields a 4–6\% gain in R1@m across benchmarks relative to vanilla GRPO, underscoring the effectiveness of capturing temporal directionality for grounding performance.  
2) Both weight adjustment and difficulty filtering contribute to consistent gains. Weight adjustment guides the model to focus on harder samples, while filtering removes already-solved examples, preventing wasted capacity. Together, these strategies deliver measurable improvements in precision and stability.  
3) Beyond component-level refinements, adopting GRPO itself provides a strong performance boost: the base VLM achieves a $10$–$30\%$ improvement on GEV benchmarks, confirming the necessity of reinforcement learning for event grounding.  

\begin{table}[t]
\centering
\setlength{\belowcaptionskip}{3pt}%
\resizebox{1.0\columnwidth}{!}{
\begin{tabular}{l|ccc}
\toprule
Method & Charades-STA & ActivityNet & TVGBench\\
\midrule
Time-R1-3B & 40.7 & 23.7 & 19.8 \\\midrule
Qwen-2.5-VL-3B & 28.7 & 15.7 & 12.6 \\
\ours-3B & \textbf{42.3} & \textbf{24.3} & \textbf{20.3} \\
\bottomrule
\end{tabular}}
\caption{mIoU metric across three GEV benchmarks.}
\label{tab:comp_GEV_miou}
\vspace{-10pt}
\end{table}

\subsection{Performance on Different VLMs}
To evaluate the versatility of our learning framework, we conducted experiments using different VLMs. Specifically, we examined the performance of our framework on Qwen-2.5-VL-Instruct-3B, and compared it with the best-performing RL baseline. Table~\ref{tab:comp_GEV_miou} presents the mIoU metric on the benchmarks. The results indicate that \ours consistently improves the performance of the base model and outperforms the best-performing baseline, indicating the effectiveness of our framework across different pre-trained VLMs. We report the performance on all metrics in Appendix~\ref{sec: additional}.

\begin{figure*}[t]   
\centering

\includegraphics[width=1.0\textwidth]{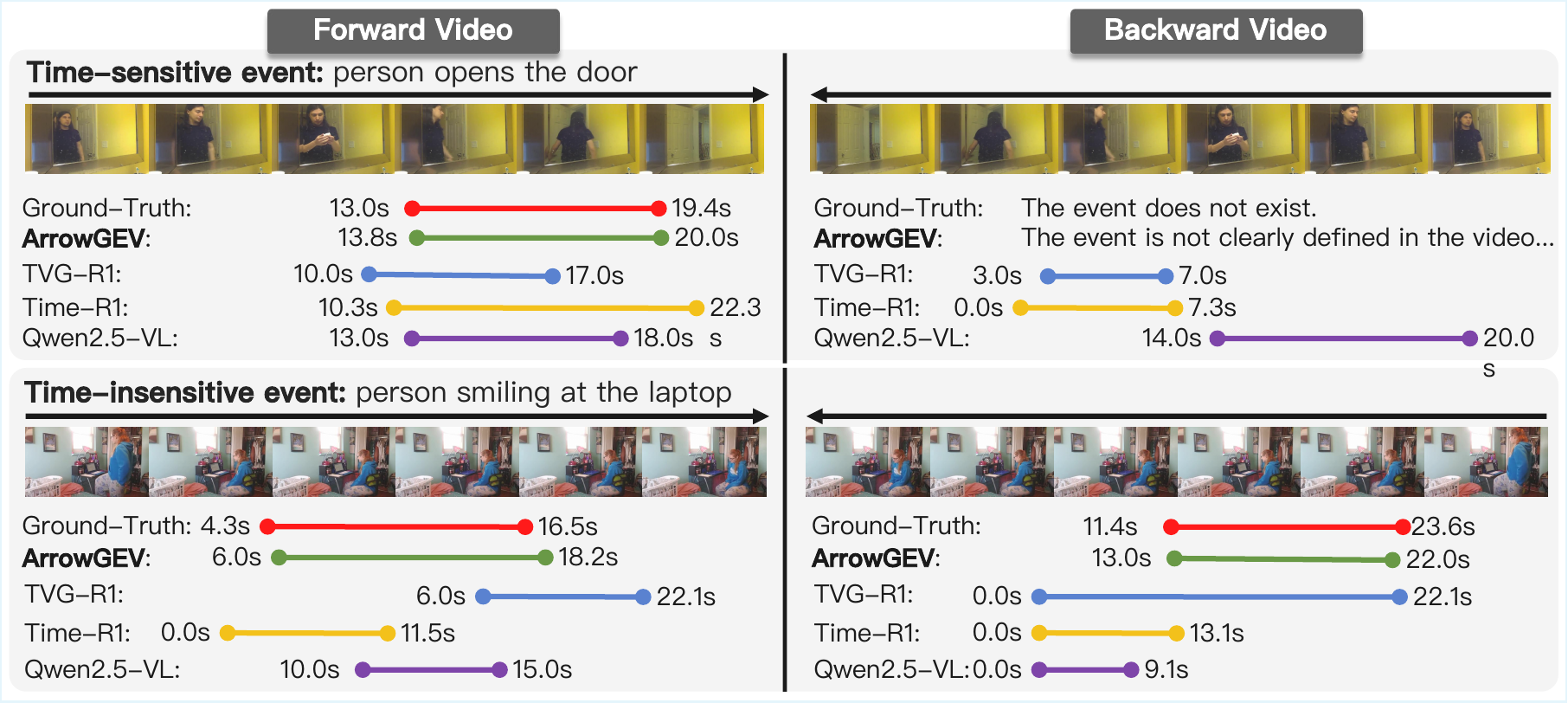}

\caption{Case study on time-sensitive events (upper row) and time-insensitive events (bottom row). Although baseline methods can partially localize the correct timestamps in the forward video, they fail to distinguish the absence of time-sensitive events in the backward video and fail to consistently localize time-insensitive events.}

\label{fig: case_study}
\vspace{-10pt}
\end{figure*}

\section{Case Study}
\label{sec: case_study}
Figure~\ref{fig: case_study} presents qualitative comparisons against high-performing RL baselines and Qwen2.5-VL-7B, evaluating both time-sensitive and time-insensitive events. For the time-sensitive event "person opens the door," \ours achieves precise localization in the forward video. Crucially, when the video is reversed, the event in the video changes to "person closes the door". \ours correctly identifies the absence of the queried event. In contrast, competing methods erroneously localize the reversed "close" action as "open", failing to comprehend temporal directionality. When analyzing the time-insensitive event "person smiling at the laptop," \ours identifies the time interval in both forward and backward videos. These results demonstrate \ours's superior understanding of temporal semantics and robustness compared to other approaches.

\section{Conclusion}
We introduce \ours, which trains VLMs to enable both event grounding and video understanding through learning temporal directionality. In addition to learning event localization in forward videos, \ours aims to recognize the absence of time-sensitive events and recognize time-insensitive events in backward videos. 
Extensive experiments demonstrate that \ours outperforms various baselines on the event grounding task and improves general video understanding and reasoning performance compared to the base model. 

\section*{Limitation}
While our method demonstrates strong performance, the computational resources required by this model are substantial. Additionally, although our work primarily focuses on grounding events in videos, we believe that learning the \textit{Arrow of Time} can be extended to more video reasoning tasks. We leave the exploration of this direction as future work.

\bibliography{custom}

\appendix

\newpage

\section*{Appendix}
\begin{table*}[t]
\centering

\setlength{\belowcaptionskip}{3pt}%
\resizebox{\textwidth}{!}{
\begin{tabular}{l|ccccc@{}|ccccc@{}|cccc}
\toprule
\multirow{2}{*}{Method} & \multicolumn{4}{c}{Charades-STA} && \multicolumn{4}{c}{ActivityNet} && \multicolumn{4}{c}{TVGBench}\\
& {\fontsize{8.4}{10}\selectfont R1@0.3} & {\fontsize{8.4}{10}\selectfont R1@0.5} & {\fontsize{8.4}{10}\selectfont R1@0.7} & {\fontsize{8.4}{10}\selectfont mIOU} && {\fontsize{8.4}{10}\selectfont R1@0.3} & {\fontsize{8.4}{10}\selectfont R1@0.5} & {\fontsize{8.4}{10}\selectfont R1@0.7} & {\fontsize{8.4}{10}\selectfont mIOU} && {\fontsize{8.4}{10}\selectfont R1@0.3} & {\fontsize{8.4}{10}\selectfont R1@0.5} & {\fontsize{8.4}{10}\selectfont R1@0.7} & {\fontsize{8.4}{10}\selectfont mIOU}\\
\midrule
Time-R1-3B & 62.6 & 40.0  & 18.2  & 40.7 && 34.8 & 19.8 &	9.0 &	23.7 &&	30.5 &	17.0 & \textbf{8.0} &	19.8 \\
Qwen-2.5-VL-3B & 42.9 & 26.9 & 12.8 & 28.7 && 22.0 & 11.5 & 4.8 & 15.7 && 18.0 & 9.8 & 5.2 & 12.6\\
\ours-3B & \textbf{64.9} & \textbf{42.5} & \textbf{19.7} & \textbf{42.3} && \textbf{36.1} & 19.4& \textbf{9.9}	& \textbf{24.3} && \textbf{31.0} & \textbf{18.6} & 7.5 & \textbf{20.3}\\
\bottomrule
\end{tabular}}
\caption{
Results on three GEV benchmarks with 3B model.
}
\label{tab: 3B}
\vspace{-10pt}
\end{table*}
\section{Event Categorization Analysis}
\label{sec: classification}
We use Qwen2.5-72B-Instruct~\citep{yang2024qwen2} to categorize the type of events. To verify the reliability of the categorization, we watch the video and label 100 events from each dataset and compute the accuracy of the model prediction. The results in Table~\ref{tab: label study} indicate that LLM's prediction is highly aligned with humans, and can well recognize whether an event is time-sensitive or time-insensitive.

\begin{table}[t]
    \centering
    \resizebox{1.0\columnwidth}{!}{%
    \begin{tabular}{lccc}
        \toprule
        \textbf{Datasets} & \textbf{Charades-STA} & \textbf{ActivityNet} & \textbf{TVGBench} \\
        \midrule
        Accuracy (\%) & 94.0 & 92.0 & 96.0 \\
        \bottomrule
    \end{tabular}
    }
    \caption{Event categorization accuracy on three benchmarks.}
    \label{tab: label study}
\end{table}
\begin{table}[t]
    \centering
    \resizebox{0.8\columnwidth}{!}{%
    \begin{tabular}{lc}
        \toprule
        \textbf{Method} & \textbf{Training time per step (s)} \\
        \midrule
        Time-R1 & 75.0  \\
        ArrowGEV & 96.7  \\
        \bottomrule
    \end{tabular}
    }
    \caption{Training time comparison.}
    \label{tab: Efficiency}
\end{table}
\section{Training Details}
\label{sec: training}
We leverage the 7B and 3B models of Qwen2.5-VL~\citep{Qwen2.5-VL} series as our base model. They are trained on large-scale image and video data and show strong instruction following and reasoning abilities. During the post-training stage, we train the model for 5 epochs, and set a batch size of 128, learning rate 2e-5, number of candidate responses $G=8$, and coefficient $\lambda=0.5$, KL term $\beta=0$, temperature in weight adjust $\tau=2$. We search these hyperparameters on the validation set. The checkpoint from the final epoch is used for all evaluations. All experiments were conducted on a single node equipped with 8 $\times$ H20 GPUs. See summary in Table~\ref{tab:hyperparameters}.

\section{Additional Results.}
\label{sec: additional}
We report the performance of \ours and compare with the high-performing RL baseline and the base model in Table~\ref{tab: 3B} on the 3B model. The results show that \ours outperforms the strong RL baseline, Time-R1-3B, on most of the metrics across all three benchmarks. These results serve as evidence for the robustness and effectiveness of our framework, demonstrating its ability to generalize and enhance the temporal reasoning capabilities of different pre-trained VLMs.

\begin{table}[t]
\centering
\resizebox{1.0\columnwidth}{!}{
\begin{tabular}{ll}
\toprule
\textbf{Hyperparameter} & \textbf{Value} \\
\midrule
Algorithm & GRPO \\
Base Model & Qwen2.5-VL-7B-Instruct \\
Max Prompt Length & 8192 \\
Max Response Length & 1024 \\
KL Coefficient($\beta$) & 0 \\
Reward Coefficient($\lambda$) & 0.5 \\
Learning Rate & 2e-5 \\
Sampling Temperature & 1.0 \\
Weighted Temperature $\tau$ & 2.0 \\
Batch Size & 128 \\
Rollout Number($G$) & 8 \\
Epochs & 5 \\
\bottomrule
\end{tabular}
}
\caption{Training Hyperparameters}
\label{tab:hyperparameters}
\end{table}

\section{Training Efficiency}
To compare training efficiencies, we recorded the average time required per training step for the Qwen2.5-VL-3B model. As detailed in Table~\ref{tab: Efficiency}, ArrowGEV incurs an approximate 28\% increase in per-step training time due to the computational overhead of processing videos in both forward and backward directions.

\section{Prompt}
\label{sec: prompt}
We list the prompt for event categorization and grounding in Figure~\ref{fig:prompt-template-sens} and Figure~\ref{fig:prompt for class}.
\begin{figure*}
  \centering %
\begin{mybox}[Prompt for time sensitivity categorization]
\begin{obeylines}
You are an AI assistant specializing in the analysis of temporal properties of events.
You will be given a sentence describing an event.

Your task is to:
1.  Analyze the event described in the sentence.
2.  Determine if the event is temporally sensitive or insensitive.
3.  Output the results in a strict JSON format without any additional text or explanations.

---
\#\#\# Input
Event Sentence: \$\{sentence\}

---
\#\#\# Evaluation Criteria
Time-Sensitive (sensitive: yes): The event has a clear forward direction. If played in reverse, it describes a different, often nonsensical or opposite, event. This indicates temporal asymmetry.
\quad Example: "A person puts a picture on the wall." (Reversed: "A person takes a picture off the wall.")
\quad Example: "A glass shatters." (Reversed: "Shards of glass assemble into a whole glass.")

Time-Insensitive (sensitive: no): The event is a continuous state or a cyclical action. If played in reverse, the fundamental nature of the event does not change. This indicates temporal symmetry.
\quad Example: "A person is playing with a light switch." (Reversed: Still looks like a person playing with a light switch.)
\quad Example: "A ball is bouncing in place." (Reversed: Still a ball bouncing in place.)
---
\#\#\# Output Format
Now, please output your result below in a JSON format by filling in the placeholders in [] without any explanations:

{
  "reason": "[Briefly explain why the event is time-sensitive or time-insensitive, describing the forward and reverse action.]",
  "sensitive": "[yes/no]"
}
"""
\end{obeylines}
\end{mybox}
\caption{The instruction for LLM to categorize events into different types.}
\label{fig:prompt-template-sens}
\end{figure*}

\begin{figure*}
  \centering %
\begin{mybox}[Prompt for event grounding]
\begin{obeylines}
To accurately pinpoint the event \$\{sentence\} in the video, determine the precise time period of the event. Output your thought process within the <think> </think> tags. Then, provide the start and end times (in seconds, precise to two decimal places) in the format "start time to end time" within the <answer> </answer> tags. For example: "12.54 to 17.83".
\end{obeylines}
\end{mybox}
\caption{The instruction to ground events in video.}
\label{fig:prompt for class}
\end{figure*}

\end{document}